\documentclass[sigconf]{acmart}

\renewcommand\footnotetextcopyrightpermission[1]{} 
\usepackage{nopageno}

\usepackage{balance}

\def\BibTeX{{\rm B\kern-.05em{\sc i\kern-.025em b}\kern-.08emT\kern-.1667em\lower.7ex\hbox{E}\kern-.125emX}}

\copyrightyear{2019}
\acmYear{2019}
\setcopyright{acmcopyright}
\acmConference[CIKM '19]{The 28th ACM International Conference on Information and Knowledge Management}{November 3--7, 2019}{Beijing, China}
\acmBooktitle{The 28th ACM International Conference on Information and Knowledge Management (CIKM'19), November 3--7, 2019, Beijing, China}
\acmPrice{15.00}
\acmDOI{10.1145/3357384.3358148}
\acmISBN{978-1-4503-6976-3/19/11}

\usepackage{booktabs}
\usepackage{amsmath}
\usepackage{flexisym}
\usepackage{dcolumn}
\usepackage{kotex}
\usepackage{color}
\usepackage{hhline}
\usepackage{verbatim}
\usepackage{multirow}
\usepackage{amssymb}
\usepackage{rotating}
\usepackage{subfigure}
\usepackage{dblfloatfix}
\usepackage{enumitem}
\newcolumntype{L}[1]{>{\raggedright\let\newline\\\arraybackslash\hspace{0pt}}m{#1}}
\newcolumntype{C}[1]{>{\centering\let\newline\\\arraybackslash\hspace{0pt}}m{#1}}
\newcolumntype{R}[1]{>{\raggedleft\let\newline\\\arraybackslash\hspace{0pt}}m{#1}}

\newcommand\Tstrut{\rule{0pt}{2.5ex}}         

\begin{document}

\fancyhead{}

\title[A Compare-Aggregate Model with Latent Clustering for Answer Selection]{A Compare-Aggregate Model with Latent Clustering\\ for Answer Selection}

%

\author{Seunghyun Yoon}
\authornote{Work conducted while the author was an intern at Adobe Research.}
\email{mysmilesh@snu.ac.kr}
\orcid{0000-0002-7262-3579}
\affiliation{%
  \institution{Seoul National University}
  \city{Seoul}
  \country{Korea}
}

\author{Franck Dernoncourt}
\email{franck.dernoncourt@adobe.com}
\affiliation{%
  \institution{Adobe Research}
  \city{San Jose}
  \state{CA}
  \country{USA}
}

\author{Doo Soon Kim}
\email{dkim@adobe.com}
\affiliation{%
  \institution{Adobe Research}
  \city{San Jose}
  \state{CA}
  \country{USA}
}

\author{Trung Bui}
\email{bui@adobe.com}
\affiliation{%
  \institution{Adobe Research}
  \city{San Jose}
  \state{CA}
  \country{USA}
}

\author{Kyomin Jung}
\email{kjung@snu.ac.kr}
\affiliation{%
  \institution{Seoul National University}
  \city{Seoul}
  \country{Korea}
}

%
\renewcommand{\shortauthors}{Yoon, et al.}

%
\begin{abstract}
In this paper, we propose a novel method for a sentence-level answer-selection task that is a fundamental problem in natural language processing.
First, we explore the effect of additional information by adopting a pretrained language model to compute the vector representation of the input text and by applying transfer learning from a large-scale corpus. 
Second, we enhance the compare-aggregate model by proposing a novel latent clustering method to compute additional information within the target corpus and by changing the objective function from listwise to pointwise.
To evaluate the performance of the proposed approaches, experiments are performed with the WikiQA and TREC-QA datasets. The empirical results demonstrate the superiority of our proposed approach, which achieve state-of-the-art performance for both datasets.
\end{abstract}

%
%



%

%

%
\maketitle

\section{Introduction}
\label{sec:introduction}
Automatic question answering (QA) is a primary objective of artificial intelligence. 
Recently, research on this task has taken two major directions based on the answer span considered by the model.
The first direction (i.e., the fine-grained approach) finds an exact answer to a question within a given passage~\cite{rajpurkar2016squad}.
The second direction (i.e., the coarse-level approach) is an information retrieval (IR)-based approach that provides the most relevant sentence from a given document in response to a question.
In this study, we are interested in building a model that computes a matching score between two text inputs.
In particular, our model is designed to undertake an answer-selection task that chooses the sentence that is most relevant to the question from a list of answer candidates.
This task has been extensively investigated by researchers because it is a fundamental task that can be applied to other QA-related tasks~\cite{wang2016compare,bian2017compare,shen2017inter,tran2018context,tay2018multi,madabushi2018integrating}.
\begin{figure*}[t]
\centering
\includegraphics[width=1.95\columnwidth]{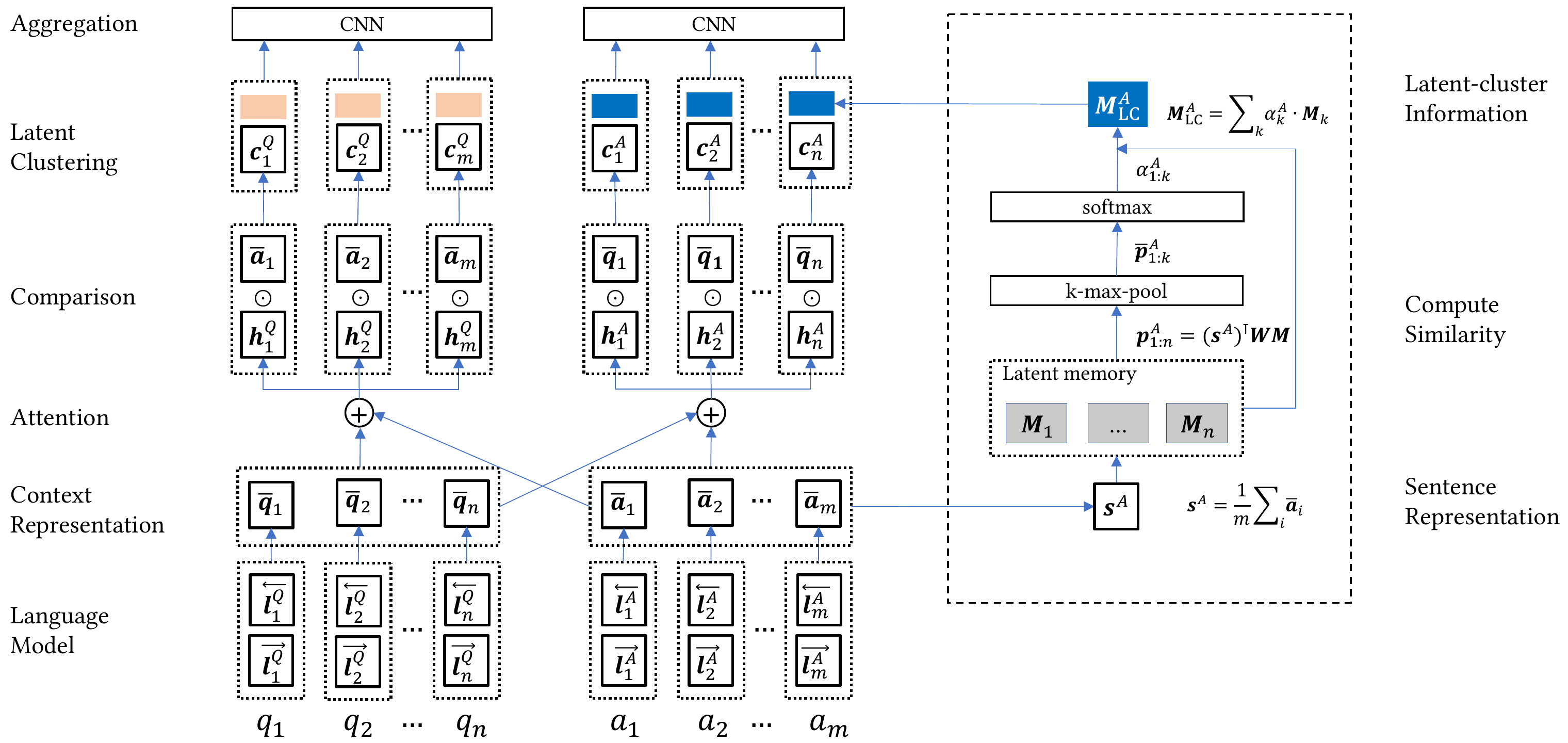}
\caption{
The architecture of the model. The dotted box on the right shows the process through which the latent-cluster information is computed and added to the answer. This process is also performed in the question part but is omitted in the figure. The latent memory is shared in both processes.
}
\label{fig:model}
\end{figure*}

However, most previous answer-selection studies have employed small datasets~\cite{wang2007jeopardy,yang2015wikiqa} compared with the large datasets employed for other natural language processing (NLP) tasks~\cite{lowe2015ubuntu,rajpurkar2016squad}.
Therefore, the exploration of sophisticated deep learning models for this task is difficult.

To fill this gap, we conduct an intensive investigation with the following directions to obtain the best performance in the answer-selection task.
First, we explore the effect of additional information by adopting a pretrained language model (\textbf{LM}) to compute the vector representation of the input text.
Recent studies have shown that replacing the word-embedding layer with a pretrained language model helps the model capture the contextual meaning of words in the sentence~\cite{peters2018deep,devlin2019bert}. 
Following this study, we select an ELMo~\cite{peters2018deep} language model for this study. 
We investigate the applicability of transfer learning (\textbf{TL}) using a large-scale corpus that is created for a relevant-sentence-selection task (i.e., question-answering NLI (QNLI) dataset~\cite{wang2018glue}).
Second, we further enhance one of the baseline models, \textbf{Comp-Clip}~\cite{bian2017compare} (refer to the discussion in \ref{ssec:compare_aggregate_model}), for the target QA task by proposing a novel latent clustering (\textbf{LC}) method.
The \textbf{LC} method computes latent cluster information for target samples by creating a latent memory space and calculating the similarity between the sample and the memory. By an end-to-end learning process with the answer-selection task, the \textbf{LC} method assigns \textit{true}-label question-answer pairs to similar clusters.
In this manner, a model will have further information for matching sentence pairs, which increases the total model performance.
Last, we explore the effect of different objective functions (listwise and pointwise learning). In contrast to previous research~\cite{bian2017compare}, we observe that the pointwise learning approach performs better than the listwise learning approach when we apply our proposed methods.
Extensive experiments are conducted to investigate the efficacy and properties of the proposed methods and show the superiority of our proposed approaches for achieving state-of-the-art performance with the WikiQA and TREC-QA datasets.

\section{Related Work}
\label{sec:realted_work}
Researchers have investigated models based on neural networks for question-answering tasks. 
One study employs a Siamese architecture that utilizes an encoder (e.g., RNN or CNN) to compute vector representations of the question and the answer. 
The affinity score is calculated based on these vector representations~\cite{lowe2015ubuntu}. 
To improve the model performance by enabling the use of information from one sentence (e.g., a question or an answer) in computing the representation of another sentence, researchers included the attention mechanism in their models~\cite{tan2015lstm,santos2016attentive,wang2017bilateral}.

Another line of research includes the compare-aggregate framework~\cite{wang2016compare}. 
In this framework, first, vector representations of each sentence are computed. Second, these representations are compared. Last, the results are aggregated to calculate the matching score between the question and the answer~\cite{bian2017compare,shen2017inter,tran2018context}. 

In this study, unlike the previous research, we employ a pretrained language model and a latent-cluster method to help the model understand the information in the question and the answer.

\section{Methods}
\label{sec:methods}

\subsection{Comp-Clip Model}
\label{ssec:compare_aggregate_model}
In this paper, we are interested in estimating the matching score $f(y|\textbf{Q},\textbf{A})$, where $y$, $\textbf{Q}\,{=}\,\{q_1, ...,q_n\}$ and $\textbf{A}\,{=}\,\{a_1, ..., a_m\}$ represent the label, the question and the answer, respectfully.
We select the model from~\cite{bian2017compare}, which is referred to as the \textbf{Comp-Clip} model, as our baseline model.
The model consists of the following four parts:

\vspace*{1mm}
\noindent\textbf{Context representation: }
The question $\textbf{Q}\,{\in}\,\mathbb{R}^{d\times Q}$ and answer $\textbf{A}\,{\in}\,\mathbb{R}^{d\times A}$, (where $d$ is a dimensionality of word embedding and Q and A are the length of the sequence in \textbf{Q} and \textbf{A}, respectively), are processed to capture the contextual information and the word as follows:
\begin{equation}
\begin{aligned}
& \overline{\textbf{Q}}=\sigma(\textbf{W}^i\textbf{Q})\odot \text{tanh}(\textbf{W}^u\textbf{Q}), \\
& \overline{\textbf{A}}=\sigma(\textbf{W}^i\textbf{A})\odot \text{tanh}(\textbf{W}^u\textbf{A}),
\end{aligned}
\label{eq:word-representation}
\end{equation}
where $\odot$ denotes element-wise multiplication, and $\sigma$ is the sigmoid function. The $\textbf{W}\,{\in}\,\mathbb{R}^{l\times d}$ is the learned model parameter.

\vspace*{1mm}
\noindent\textbf{Attention: }
The soft alignment of each element in $\overline{\textbf{Q}}\,{\in}\,\mathbb{R}^{l\times Q}$ and $\overline{\textbf{A}}\,{\in}\,\mathbb{R}^{l\times A}$ are calculated using dynamic-clip attention~\cite{bian2017compare}. We obtain the corresponding vectors $\textbf{H}^Q\,{\in}\,\mathbb{R}^{l\times A}$ and $\textbf{H}^A\,{\in}\,\mathbb{R}^{l\times Q}$.
\begin{equation}
\begin{aligned}
& \textbf{H}^Q= \overline{\textbf{Q}}\cdot~\text{softmax}({(\textbf{W}^{q}\overline{\textbf{Q}})}^{\intercal}\overline{\textbf{A}}),\\
& \textbf{H}^A= \overline{\textbf{A}}\cdot~\text{softmax}({(\textbf{W}^{a}\overline{\textbf{A}})}^{\intercal}\overline{\textbf{Q}}).
\end{aligned}
\label{eq:attention}
\end{equation}

\vspace*{1mm}
\noindent\textbf{Comparison: }
A comparison function is used to match each word in the question and answer to a corresponding attention-applied vector representation:
\begin{equation}
\begin{aligned}
& \textbf{C}^Q = \overline{\textbf{A}} \odot \textbf{H}^Q, ~~(\textbf{C}^Q\,{\in}\,\mathbb{R}^{l\times A}),\\
& \textbf{C}^A = \overline{\textbf{Q}} \odot \textbf{H}^A, ~~(\textbf{C}^A\,{\in}\,\mathbb{R}^{l\times Q}),
\end{aligned}
\label{eq:comparison}
\end{equation}
where $\odot$ denotes element-wise multiplication.

\vspace*{1mm}
\noindent\textbf{Aggregation: }
We aggregate the vectors from the comparison layer using CNN~\cite{kim2014convolutional} with $n$-types of filters and calculate the matching score between \textbf{Q} and \textbf{A}.
\begin{equation}
\begin{aligned}
& \textbf{R}^Q=\text{CNN}(\textbf{C}^Q),~~\textbf{R}^A=\text{CNN}(\textbf{C}^A), \\
& \text{score}=\sigma([\textbf{R}^Q;\textbf{R}^A]^{\intercal}\,\textbf{W}),
\end{aligned}
\label{eq:aggregation}
\end{equation}
where [;] denotes concatenation of each vector $\textbf{R}^Q\,{\in}\,\mathbb{R}^{nl}$ and $\textbf{R}^A\,{\in}\,\mathbb{R}^{nl}$. The $\textbf{W}\,{\in}\,\mathbb{R}^{2nl\times 1}$ is the learned model parameter.

\subsection{Proposed Approaches}
\label{ssec:proposed-approaches}
To achieve the best performance in the answer-selection task, we propose four approaches: adding a pretrained \textbf{LM}; adding the \textbf{LC} information of each sentence as auxiliary knowledge; applying \textbf{TL} to benefit from large-scale data; and modifying the objective function from listwise to pointwise learning. Figure~\ref{fig:model} depicts the total architecture of the proposed model.

\vspace*{2mm}
\noindent\textbf{Pretrained Language Model (LM): }
Recent studies have shown that replacing the word embedding layer with a pretrained \textbf{LM} helps the model capture the contextual meaning of the words in the sentence~\cite{peters2018deep,devlin2019bert}. 
We select an ELMo~\cite{peters2018deep} language model and replace the previous word embedding layer with the ELMo model as follows:
$\textbf{L}^{\text{Q}}\,{=}\,\text{ELMo}(\textbf{Q}),~\textbf{L}^{\text{A}}\,{=}\,\text{ELMo}(\textbf{A})$. 
These new representations—$\textbf{L}^{\text{Q}}$ and $\textbf{L}^{\text{A}}$—are substituted for \textbf{Q} and \textbf{A}, respectively, in equation~(\ref{eq:word-representation}).

\vspace*{2mm}
\noindent\textbf{Latent Clustering (LC) Method: }
We assume that extracting the \textbf{LC} information of the text and using it as auxiliary information will help the neural network model analyze the corpus.
The dotted box in figure~\ref{fig:model} shows the proposed \textbf{LC} method.
We create \textit{n}-many latent memory vectors $\textbf{M}_{1:n}$ and calculate the similarity between the sentence representation and each latent memory vector. 
The latent-cluster information of the sentence representation will be obtained using a weighted sum of the latent memory vectors according to the calculated similarity as follows:
\begin{equation}
\begin{aligned}
& \textbf{p}_{1:n} = {\textbf{s}}^{\intercal}\textbf{W}\textbf{M}_{1:n}, \\
& \overline{\textbf{p}}_{1:k} = \text{\textit{k}-max-pool}(\textbf{p}_{1:n}), \\
& \alpha_{1:k} = \text{softmax}(\overline{\textbf{p}}_{1:k}), \\
& \textbf{M}_{\text{LC}}={\scriptstyle\sum_k}\overline{\alpha}_k{\textbf{M}_k},
\end{aligned}
\label{eq:kLC}
\end{equation}
where $\textbf{s}\,{\in}\,\mathbb{R}^{d}$ is a sentence representation, $\textbf{M}_{1:n}\,{\in}\,\mathbb{R}^{d'\times n}$ indicates the latent memory, and $\textbf{W}{\in}\,\mathbb{R}^{d \times d'}$ is the learned model parameter.

We apply the \textbf{LC} method and extract cluster information from each question and answer. This additional information is added to each of the final representations in the comparison part~(see \ref{ssec:compare_aggregate_model}) as follows:
\begin{equation}
\begin{aligned}
& \textbf{M}_{\text{LC}}^Q=f(({\scriptstyle\sum_i}\overline{q}_i)/n),~\overline{q}_i\subset{\overline{\textbf{Q}}_{1:n}}, \\
& \textbf{M}_{\text{LC}}^A=f(({\scriptstyle\sum_i}\overline{a}_i)/m),~\overline{a}_i\subset{\overline{\textbf{A}}_{1:m}}, \\
& \textbf{C}^Q_{\text{new}}= [\textbf{C}^Q;\textbf{M}_{\text{LC}}^Q],~~\textbf{C}^A_{\text{new}}= [\textbf{C}^A;\textbf{M}_{\text{LC}}^A], \\
\end{aligned}
\label{eq:lc}
\end{equation}
where $f$ is the \textbf{LC} method (in equation~\ref{eq:kLC}) and [;] denotes the concatenation of each vector.
These new representations—$\textbf{C}^Q_{\text{new}}$ and $\textbf{C}^A_{\text{new}}$—are substituted for $\textbf{C}^Q$ and $\textbf{C}^A$ in equation~(\ref{eq:aggregation}).
Note that we average word-embedding to obtain sentence representation in the previous equation.

\vspace*{2mm}
\noindent\textbf{Transfer Learning (TL): }
To observe the efficacy in a large dataset, we apply transfer learning using the question-answering NLI (QNLI) corpus~\cite{wang2018glue}.
We train the \textbf{CompClip} model with the QNLI corpus and then fine-tune the model with target corpora, such as the WikiQA and TREC-QA datasets.

\vspace*{2mm}
\noindent\textbf{Pointwise Learning to Rank: }
Previous research adopts a listwise learning approach.
With a dataset that consists of a question, \textbf{Q}, a related answer set, $\textbf{A}\,{=}\,\{\textbf{A}_1, ..., \textbf{A}_N\}$, and a target label, $\textbf{y}\,{=}\,\{y_1, ..., y_N\}$, a matching score is computed using equation (\ref{eq:aggregation}).
This approach applies KL-divergence loss to train the model as follows:
\begin{equation}
\begin{aligned}
& \text{score}_i = \text{model}(\textbf{Q}, \textbf{A}_i), \\
& \textbf{S}=\text{softmax}([\text{score}_1, ..., \text{score}_i]), \\
& \text{loss} = {\scriptstyle\sum_{n=1}^{N}}\text{KL}(\textbf{S}_n||\textbf{y}_n),
\end{aligned}
\label{eq:pointw-wise-learning-to-rank}
\end{equation}
where $i$ is the number of answer candidates for the given question and $N$ is the total number of samples employed during training.


In contrast, we pair each answer candidate to the question and compute the cross-entropy loss to train the model as follows:
\begin{equation}
\begin{aligned}
& \text{loss} = -{\scriptstyle\sum_{n=1}^{N}}~{y}_n~\text{log}~(\text{score}_n),
\end{aligned}
\label{eq:point-wise-loss}
\end{equation}
where $N$ is the total number of samples used during training. 
Using this approach, the number of training instances for a single iteration increases, as shown in table~\ref{table:stat}.

\begin{table}[t]
\centering
\caption{
Properties of the dataset.
}
\begin{tabular}{C{0.18\columnwidth}C{0.08\columnwidth}C{0.08\columnwidth}C{0.08\columnwidth}C{0.08\columnwidth}C{0.08\columnwidth}C{0.08\columnwidth}}

\toprule
\multirow{2}{*}{\textbf{Dataset}} & \multicolumn{3}{c}{\textbf{Listwise pairs}} & \multicolumn{3}{c}{\textbf{Pointwise pairs}} \\
\cmidrule{2-7}
 	& train & dev & test & train & dev & test \\
\midrule
WikiQA\Tstrut & 873\Tstrut & 126\Tstrut & 243\Tstrut & 8.6k\Tstrut & 1.1k\Tstrut & 2.3k\Tstrut \\
TREC-QA & 1.2k & 65 & 68 & 53k & 1.1k & 1.4k \\
QNLI & 86k & 10k & - & 428k & 169k & - \\

\bottomrule
\end{tabular}
\label{table:stat}
\end{table}
\begin{table*}[t]
\centering
\caption{
Model performance (the top 3 scores are marked in bold for each task). 
We evaluate model~\cite{wang2016compare,bian2017compare,shen2017inter,tran2018context} on the WikiQA corpus using author’s implementation (marked by *).
For TREC-QA case, we present reported results in the original papers.
}
\begin{tabular}{L{0.8\columnwidth}|C{0.1\columnwidth}C{0.1\columnwidth}C{0.1\columnwidth}C{0.1\columnwidth}|C{0.1\columnwidth}C{0.1\columnwidth}C{0.1\columnwidth}C{0.1\columnwidth}}
\toprule
\multirow{3}{*}{\textbf{Model}}   & \multicolumn{4}{c|}{\textbf{WikiQA}}  & \multicolumn{4}{c}{\textbf{TREC-QA}} \\ 
\cline{2-9}              
& \multicolumn{2}{c}{MAP}\Tstrut   & \multicolumn{2}{c|}{MRR}\Tstrut   & \multicolumn{2}{c}{MAP}\Tstrut   & \multicolumn{2}{c}{MRR}\Tstrut \\
\cline{2-9}              
& dev\Tstrut & test\Tstrut & dev\Tstrut & test\Tstrut & dev\Tstrut & test\Tstrut & dev\Tstrut & test\Tstrut \\
\hline
Compare-Aggregate (2017)~\cite{wang2016compare}\Tstrut & 0.743*\Tstrut & 0.699*\Tstrut & 0.754*\Tstrut & 0.708*\Tstrut &
-\Tstrut & -\Tstrut & -\Tstrut & -\Tstrut\\
Comp-Clip (2017)~\cite{bian2017compare}         & 0.732* & 0.718* & 0.738* & 0.732* & - & 0.821 & - & 0.899\\
IWAN (2017)~\cite{shen2017inter}              & 0.738* & 0.692* & 0.749* & 0.705* & - & 0.822 & - & 0.899\\
IWAN + sCARNN (2018)~\cite{tran2018context}     &
0.719* & 0.716* & 0.729* & 0.722* &
 - & 0.829 & - & 0.875\\
MCAN (2018)~\cite{tay2018multi}     & - & - & - & - & - & 0.838 & - & \textbf{0.904}\\
Question Classification (2018)~\cite{madabushi2018integrating}     & - & - & - & - & - & \textbf{0.865} & - & 0.904\\
\hline
\multicolumn{9}{l}{\textbf{Listwise Learning to Rank}}\Tstrut \\
\hline
Comp-Clip (our implementation)\Tstrut               & 0.756\Tstrut & 0.708\Tstrut & 0.766\Tstrut & 0.725\Tstrut & 0.750\Tstrut & 0.744\Tstrut & 0.805\Tstrut & 0.791\Tstrut\\
Comp-Clip (our implementation) + LM          & 0.783 & 0.748 & 0.791 & 0.768 & 0.825 & 0.823 & 0.870 & 0.868\\
Comp-Clip (our implementation) + LM + LC     & 0.787 & 0.759 & 0.793 & 0.772 & 0.841 & 0.832 & 0.877 & 0.880\\
Comp-Clip (our implementation) + LM + LC +TL & 0.822 & \textbf{0.830} & 0.836 & \textbf{0.841} & 0.866 & 0.848 & 0.911 & 0.902\\
\hline
\multicolumn{9}{l}{\textbf{Pointwise Learning to Rank}}\Tstrut \\
\hline
Comp-Clip (our implementation)\Tstrut               & 0.776\Tstrut & 0.714\Tstrut & 0.784\Tstrut & 0.732\Tstrut & 0.866\Tstrut & 0.835\Tstrut & 0.933\Tstrut & 0.877\Tstrut\\
Comp-Clip (our implementation) + LM          & 0.785 & 0.746 & 0.789 & 0.762 & 0.872 & 0.850 & 0.930 & 0.898\\
Comp-Clip (our implementation) + LM + LC     & 0.782 & \textbf{0.764} & 0.785 & \textbf{0.784} & 0.879 & \textbf{0.868} & 0.942 & \textbf{0.928}\\
Comp-Clip (our implementation) + LM + LC +TL & 0.842 & \textbf{0.834} & 0.845 & \textbf{0.848} & 0.913 & \textbf{0.875} & 0.977 & \textbf{0.940}\\
\bottomrule
\end{tabular}
\label{table:wiki_trec}
\end{table*}

\section{Empirical Results}
\label{empirical_results}
We regard all tasks as relevant answer selections for the given questions.
Following the previous study, we report the model performance as the mean average precision (MAP) and the mean reciprocal rank (MRR)\footnote{\url{https://aclweb.org/aclwiki/Question_Answering_(State_of_the_art)}}.
To test the performance of the model, we utilize the TREC-QA, WikiQA and QNLI datasets~\cite{wang2007jeopardy, yang2015wikiqa, wang2018glue}.

\subsection{Dataset}
\label{sec:dataset}
\vspace*{2mm}
\noindent\textbf{WikiQA}
\cite{yang2015wikiqa} is an answer selection QA dataset constructed from real queries of Bing and Wikipedia. Following the literature~\cite{bian2017compare,shen2017inter}, we use only questions that contain at least one correct answer among the list of answer candidates. There are 873/126/243 questions and 8,627/1,130/2,351 question-answer pairs for train/dev/test split.

\vspace*{2mm}
\noindent\textbf{TREC-QA}
\cite{wang2007jeopardy} is another answer selection QA dataset created from the TREC Question-Answering tracks. In this study, we use the clean dataset that removed questions from the dev and test datasets that did not have answers or had only positive/negative answers. There are 1,229/65/68 questions and 53,417/1,117/1,442 question-answer pairs for train/dev/test split.

\vspace*{2mm}
\noindent\textbf{QNLI}
\cite{wang2018glue} is a modified version of the SQuAD dataset~\cite{rajpurkar2016squad} that allows for sentence selection QA. The context paragraph in SQuAD is split into sentences, and each sentence is paired with the question. The true label is given to the question-sentence pairs when the sentence contains the answer. There are 86,308/10,385 questions and 428,998/169,435 question-answer pairs for train/dev split. Considering the large size of this dataset, we use it to train the base model for transfer learning; it is also used to evaluate the proposed model performance in a large dataset environment.

\subsection{Implementation Details}
\label{sec:implementation_details}
To implement the \textbf{Comp-Clip} model, we apply a context projection weight matrix with 100 dimensions that are shared between the question part and the answer part (eq.~\ref{eq:word-representation}). In the aggregation part, we use 1-D CNN with a total of 500 filters, which involves five types of filters $K\,{\in}\,\mathbb{R}^{\{1,2,3,4,5\}\times100}$, 100 per type. 
This CNN is independently applied to the question part and answer part.
For the \textbf{LC} method, we perform additional hyper-parameter searching experiments to select the best parameters. We select \textit{k} (for the \textit{k}-max-pool in equation~\ref{eq:kLC}) as 6 and 4 for the WikiQA and TREC-QA case, respectively. In both datasets, we apply 8 latent clusters.

The vocabulary size in the WiKiQA, TREC-QA and QNLI dataset are 30,104, 56,908 and 154,442, respectively. 
When applying the \textbf{TL}, the vocabulary size is set to 154,442, and the dimension of the context projection weight matrix is set to 300.
We use the Adam optimizer, including gradient clipping, by the norm at a threshold of 5.
For the purpose of regularization, we applied a dropout with a ratio of 0.5.

\subsection{Comparison with Other Methods}
\label{ssec:comparison-with-other-methods}
Table~\ref{table:wiki_trec} shows the model performance for the WikiQA and TREC-QA datasets.
For the Compare-Aggregate (2016), Comp-Clip (2017), IWAN (2017) and IWAN+sCARNN (2018) models, we measure the performance on the WikiQA dataset using the authors' implementations (marked by * in the table).
Unlike previous studies, we report our results for both the dev dataset and the test dataset because we note a performance gap between these datasets. While training the model, we apply an early stop that is based on the performance of the dev dataset and measure the performance on the test dataset.
Because \textbf{Comp-Clip}~\cite{bian2017compare} is our baseline model, we implement it from scratch and achieve a performance that is similar to that of the original paper.

\vspace*{1mm}
\noindent\textbf{WikiQA: }
For the WikiQA dataset, the pointwise learning approach shows a better performance than the listwise learning approach.
We combine \textbf{LM} with the base model (\textbf{Comp-Clip} +\textbf{LM}) and observe a significant improvement in performance in terms of MAP (0.714 to 0.746 absolute). 
When we add the \textbf{LC} method (\textbf{Comp-Clip} +\textbf{LM} +\textbf{LC}), the best previous results are surpassed in terms of MAP (0.718 to 0.764 absolute). We achieve a vast improvement in performance in terms of the MAP (0.764 to 0.834 absolute) by including the \textbf{TL} approach (\textbf{Comp-Clip} + \textbf{LM} + \textbf{LC} + \textbf{TL}).

\vspace*{1mm}
\noindent\textbf{TREC-QA: }
The pointwise learning approach also shows excellent performance with the TREC-QA dataset.
As shown in table 1, the TREC-QA dataset has a larger number of answer candidates per question. We assume that this characteristic prevents the model from handling the dataset with a listwise learning approach.
As in the WikiQA case, we achieve additional performance gains in terms of the MAP as we apply \textbf{LM}, \textbf{LC}, and \textbf{TL} (0.850, 0.868 and 0.875, respectively).
In particular, our model outperforms the best previous result when we add \textbf{LC} method, (\textbf{Comp-Clip}
+\textbf{LM} +\textbf{LC}) in terms of MAP (0.865 to 0.868).

\begin{table}[t]
\centering
\caption{
Model (Comp-Clip +LM +LC) performance on the QNLI corpus with a variant number of clusters (top score marked as bold).
}
\begin{tabular}{C{0.20\columnwidth}C{0.14\columnwidth}C{0.14\columnwidth}C{0.14\columnwidth}C{0.14\columnwidth}}
\toprule
\multirow{2}{*}{\textbf{\# Clusters}} & \multicolumn{2}{c}{\textbf{Listwise Learning}} & \multicolumn{2}{c}{\textbf{Pointwise Learning}} \\
\cmidrule{2-5}
& MAP & MRR & MAP & MRR \\
\midrule
1  & 0.822 & 0.819 & 0.842 & 0.841\\
4  & 0.839 & 0.840 & 0.846 & 0.845\\
8  & \textbf{0.841} & \textbf{0.842} & 0.846 & \textbf{0.846}\\
16 & 0.840 & \textbf{0.842} & \textbf{0.847} & \textbf{0.846}\\
\bottomrule
\end{tabular}
\label{table:num_cluster}
\end{table}

\subsection{Impact of Latent Clustering}
\label{sec:qnli_evaluation}
To evaluate the impact of latent clustering method (\textbf{Comp-Clip} +\textbf{LM} +\textbf{LC}) in a larger dataset environment, we perform QNLI evaluation.
Table~\ref{table:num_cluster} shows the performance of the model (\textbf{Comp-Clip} +\textbf{LM} +\textbf{LC}) for the QNLI dataset with a variant number of clusters.
Note that the QNLI dataset is created from the SQuAD~\cite{rajpurkar2016squad} dataset, which only provides train and dev subsets.
Consequently, we report the model performances for the dev dataset.
As shown in the table, we achieve the best results with 8 clusters in listwise learning and 16 clusters in pointwise learning. In both cases, we achieve no additional performance gain after 16 clusters.

\section{Conclusion}
\label{sec:conclusion}
In this study, our proposed method achieves state-of-the-art performance for both the WikiQA dataset and TREC-QA dataset. We show that leveraging a large amount of data is crucial for capturing the contextual representation of input text. In addition, we show that the proposed latent clustering method with a pointwise objective function significantly improves the model performance in the sentence-level QA task.

\begin{acks}
We sincerely thank Carl I. Dockhorn and Yu Gong at Adobe for their in-depth feedback for this research. 
K. Jung is with ASRI, Seoul National University, Korea.
This work was supported by the Ministry of Trade, Industry \& Energy (MOTIE, Korea) under Industrial Technology Innovation Program (No.10073144) and by the NRF funded by the Korea government (MSIT) (No. 2016M3C4A7952632).
\end{acks}

%

%
\bibliographystyle{ACM-Reference-Format}
\bibliography{arXiv}

%

\end{document}